\documentclass[journal]{IEEEtran}
\usepackage{graphicx}
\usepackage{xcolor}
\usepackage{multirow}
\usepackage{amsmath}

\ifCLASSINFOpdf
\else
\fi

\usepackage{epstopdf}
\begin{document}
%
\title{Unsupervised Learning Consensus Model for  Dynamic Texture Videos Segmentation}
%
%
%

\author{Lazhar~Khelifi,
        and~Max~Mignotte
\thanks{L. Khelifi and M. Mignotte are with Department of Computer Science and Operations Research, Montreal University, Montreal, Quebec, Canada (e-mail: khelifil@iro.umontreal.ca, mignotte@iro.umontreal.ca. (see http://www.lazharkhelifi.com)}
}

\maketitle

\begin{abstract}
	Dynamic texture (DT) segmentation, and video processing in general, is currently widely dominated by methods based on deep neural networks that require the deployment of a large number of layers. Although this parametric approach has shown superior performances for  the dynamic texture segmentation,
	all current deep learning methods suffer from a
	significant main weakness related to the lack of a sufficient
	reference annotation to train models and to make them functional.
	In addition, the result of these methods can deteriorate
	significantly when the network is fed  with images or video  not similar (as regards, shape, texture, color, etc.) to the images previously included in the training dataset. This study explores the
	unsupervised segmentation approach  that can be used in the absence
	of training data to segment new videos. In particular, it tackles
	the task of dynamic texture segmentation. By automatically assigning a single class label to each region or group, this task consists of clustering into groups complex phenomena and characteristics which are both spatially and temporally repetitive. We present an effective unsupervised learning consensus model  for the segmentation of dynamic texture (ULCM). This model is designed to merge different segmentation maps that contain multiple and weak quality regions in order to achieve a more accurate final result of segmentation. The diverse labeling fields required for the combination process are obtained by a simplified grouping scheme applied to an input video (on the basis of a three orthogonal planes: $xy$, $yt$ and $xt$). 
	In the proposed model, the set of values of the requantized local binary patterns (LBP) histogram around the pixel to be classified are used as features which represent  both the spatial and temporal information replicated in the video.
	Experiments conducted on  the challenging SynthDB dataset  show that, contrary to current dynamic texture segmentation approaches that either require parameter estimation or a training step, ULCM is significantly faster, easier to code, simple and has limited parameters.  Further qualitative experiments based on the YUP++ dataset prove the efficiently and  competitively of the ULCM.
\end{abstract}


%
\IEEEpeerreviewmaketitle

\section{Introduction}
\label{Introduction}

\begin{figure}
			\includegraphics[width=8.5cm]{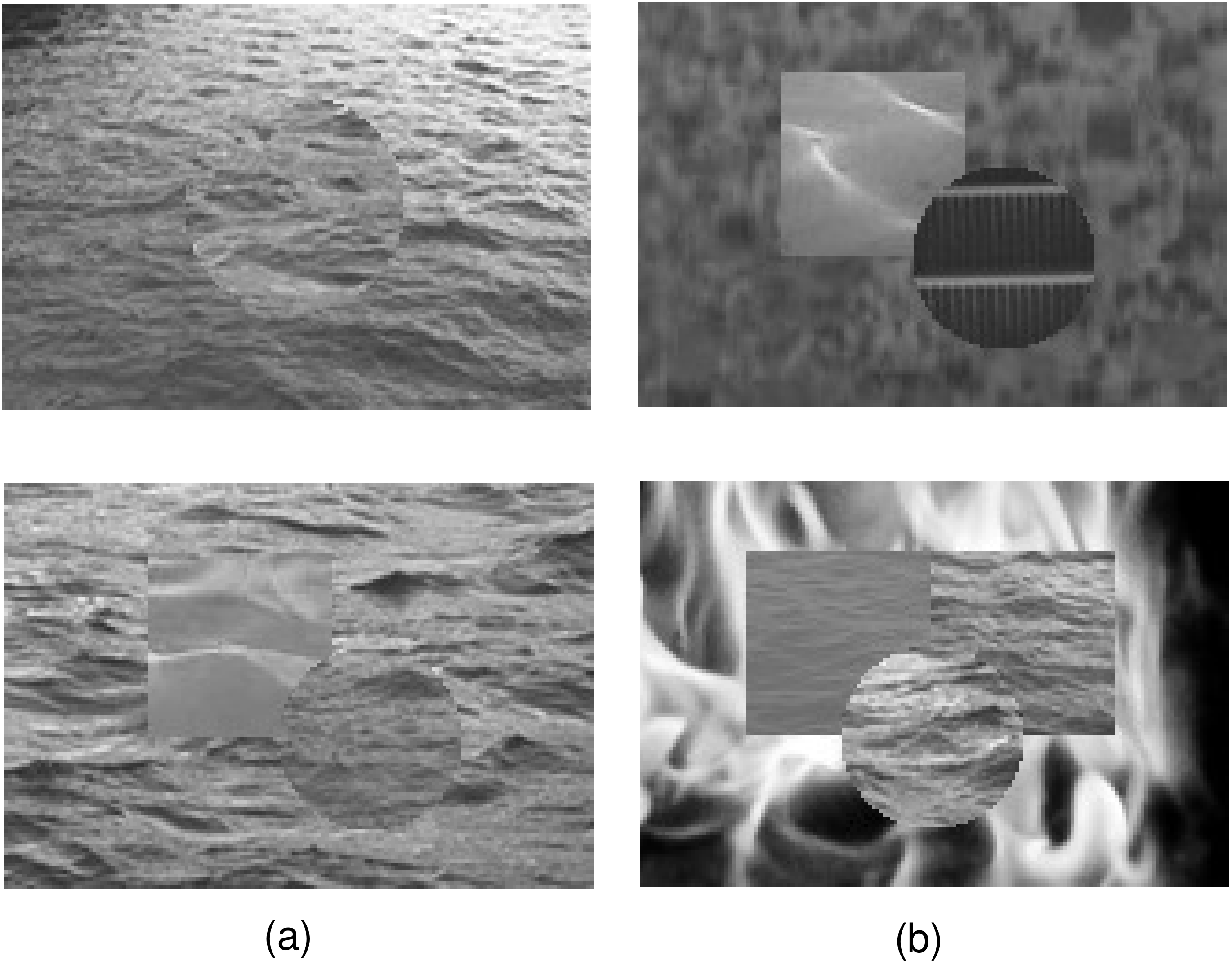}
			\caption{ \label{Fig1} Examples of DTs. (a) DTs are different regarding motion or displacement (i.e. in temporal mode), but similar in terms of appearance (i.e. in spatial mode) related mainly to texture. (b) DTs are different with respect to appearance, but similar in terms of motion $^{1}$ .}
		\end{figure}

	\IEEEPARstart{D}{ynamic texture}, or texture movie, combines texture in
	the spatial  domain  with  motion (with some form of stationarity) in
	the temporal  domain \cite{Hajati2017} (see Fig. \ref{Fig1}).
	Consequently, dynamic texture segmentation can be very complex because
	this process requires to jointly analyze spatiotemporal data which can
	be very different in nature, just like the numerous dynamic scenes
	existing in the real world, such as; cloud, falling snow, flowing
	flag, swirl, smoke, etc. \cite{Chen2013}.

		%
	
	\footnotetext[1]{	http://www.ee.oulu.fi/$\sim$gyzhao/research/dynamic$\_$texture$\_$recognition.htm}

	\vspace{1ex}
	Recently, research on dynamic segmentation of textures has been of
	growing interest and has led to the development of interesting and
	varied methods.
	Doretto \textit{et al.} \cite{Doretto2003}  used spatio-temporal
	statistics and more precisely their dynamics over time with
	Gauss-Markov models \cite{Sun2013} to segment a sequence of images into regions. A
	variational optimization framework was then used to infer the
	parameters of the model and to locate the boundary of each region.
	However, a limitation of this model is based on the assumption that
	regions very slowly over time and also essentially according to the
	irradiance within each region.
	Vidal \textit{et al.}\cite{Vidal2005}, for their part, tackled this
	problem by first analyzing a generalized principal component analysis
	(GPCA) of the optical flow field generated from the video which was finally
	exploited to segment the spatiotemporal data by grouping pixels having
	similar trajectories in time. Nevertheless, as it was originally
	designed, this segmentation model is limited to only two classes.
	Chan \textit{et al.} \cite{Chan2008} proposed the mixture of dynamic textures (DTM) as  an appropriate representation for both the dynamics  and   appearance of dynamic texture videos.  In their model, the different parameters are learned using  the expectation-maximization (EM) algorithm. Their work was extended in \cite{Fernández2018}  by
	using the efficiency of the GPU computations to accelerate the
	segmentation process.
	Wattanachote \textit{et al.} \cite{Wattanachote2016} presented a new
	and original method of semi-automatic dynamic segmentation of textures by exploiting motion vectors generated from the model of Farneb$\ddot{a}$ck\footnotemark[2] \cite{Farneback2003}. 
	Nevertheless, an important constraint of this technique is that user interaction is still required to select the focus objects and to fine-tune the result to produce a high quality spatiotemporal segmentation map.
	%
	\footnotetext[2]{An algorithm for estimating dense optical flow based
		on modeling the neighborhoods of each pixel by quadratic polynomials.}
	%
	Nguyen \textit{et al.} \cite{Nguyen2014} proposed a novel automatic
	feature selection dynamic mixture model (FSDTM) to solve the motion
	segmentation problem.  
	The key strength of their approach is that it is totally unsupervised and does not require a set of training data
	having known classifications on which to fit the mixture model. In
	this approach, the expectation maximization (EM) method is
	exploited to estimate the parameters of the  model of mixture in the
	maximum likelihood sense.  However, the EM algorithm remains very
	sensitive to initial values, noise, outliers and to the shapes of the
	laws of distribution chosen a priori in the mixture model and has also
	the drawback of converging at local minima.
	An interesting (but partially supervised) approach combining a supervised learning method with a filter-based motion features  has
	been introduced by Teney \textit{et al.}  \cite{Teney2015}.

	\vspace{1ex}
	Different from the existing methods,  Cai  \textit{et al.}
	\cite{Cai2019}  have suggested a new dynamic texture methodology for
	ultrasound images. This model consists in a combination  of surfacelet transform, parallel computing and HMT
	model. One advantage of this work is that makes it possible to build a model that   both cover temporal and spatial  information by considering simultaneously a sequence of images. Yousefi \textit{et al.} \cite{Yousefi2019} proposed an interesting
	non-parametric fully Bayesian approach for DT segmentation, built based on
	joining Dirichlet process mixture (DPM)  with
	generative dynamic texture models (GDTMs). This method effectively eliminates the required information on the amount of textures and its initial partitioning. In \cite{Paygude2018} authors
	discussed three DT segmentation methods  based on  global
	spatiotemporal technique (contourlet transform), local
	spatiotemporal technique (local binary pattern) and optical flow. Their experimentation  conducted through these individual techniques and also using  certain combinations of them. Results showed that local binary pattern is simple  to be implemented , less computationally  complex and  represents a suitable variant can be considered depending on the application at hand. However, optical flow technique is more computationally  complex but still represents a natural way of motion detection. This study also showed  the capability of  contourlet transform in tracing smooth contours, especially, in case of images that contain natural DTs.
	Among the most recent work, one can cite the algorithm proposed by
	Andrearczyk \textit{et al.}  \cite{Andrearczyk2017} in which  a convolutional neural networks (CNNs)  is applied on three orthogonal slices $xt$, $xy$ and $yt$ of an input video sequence.  The major drawback in their approach is that the training  of independent CNNs on three orthogonal planes, and the
	combination of their outputs makes  the process more complex from a
	computational point of view while being also supervised.  
	The major drawback  of their approach is that the training of individual CNNs on three orthogonal planes, as well as  the combination of their outputs, render the entire procedure more computationally complex from a computational point of view while being also supervised.  
	
	\vspace{1ex}
	Motivated by the above observations, we herein introduce a novel fusion
	model for dynamic texture segmentation called ULCM. Our model combines  multiple and soft segmentation maps in order to obtain a
	more consistent and high-quality spatiotemporal segmentation result.   
	These initial and weak partition maps are estimated from
	separate slices (or frames) of the video sequence and across
	the different axis of the data cube.  In addition, in order to overcome
	the disadvantages of previous methods that often lead to complex
	estimation, optimization or combinatorial problems, we herein propose
	a simple energy-based model based on an efficient segmentation fusion
	criterion  derived from the global consistency error  (GCE).  This metric of GCE is a perceptual measure that considers 
	the inherently multi-scale property of any image partition (potentially qualified as a refinement of an existing segmentation) by quantifying the level of difference of two segmentation maps.  Moreover,
	to efficiently optimizing our energy-based model, we introduce a modified local optimization scheme derived from the Iterative Conditional Mode method  (ICM).

	\vspace{1ex}
	In summary, this study provides the following three main contributions:  
	%
	\begin{itemize}  
		\item  
		A new consensus model of unsupervised learning is proposed for the dynamic segmentation of textures. The developed model combines multiple and weak segmentation results to obtain a finer and more reliable segmentation of an input video.
		\item An energy function derived from the global coherence error (GCE) is proposed for the fusion process. The GCE measure is a perceptual criterion that considers the inherently multi-scaled nature of an image segmentation (by calculating the degree of refinement within two spatial segments).
		\item 
		Extensive experiments on two reference datasets demonstrate the effectiveness of the proposed unsupervised method and its ability to achieve high quality segmentation results with clear boundaries.   
	\end{itemize}
	
	\vspace{1ex}
	The rest of the article is arranged as follows: First of all, we provide
	a brief definition of dynamic texture problem in section
	\ref{Texture}. In Section \ref{Method}, we introduce the  proposed ULCM
	fusion model. In Section \ref{Experiments}, we present an experimental
	evaluation of the developed algorithm using synthetic and real video
	datasets. Finally, in Section \ref{Conclusion}, we draw a conclusion.
	
	%
	\section{Dynamic Texture}
	\label{Texture}
	While a variety of definitions of the dynamic texture has been
	suggested, this paper will use the definition first suggested by Chan
	\textit{et al.}  \cite{Chan2008} who define it as a generative model
	for both the appearance (frame of video  at time $t$), and the dynamics of
	video sequences (temporal evolution of the video), based on a linear
	dynamic system. Following this definition, a linear function of the current state vector, plus some observation noise, generally represents the appearance of the image $y_{t} \in R^{n}$, while the state process evolving over time $x_{t} \in R^{n}$ (typically $n \ll m$) represents the dynamics. Mathematically, the
	equations of this system are defined as follows:
	
	%
	\begin{eqnarray}
	\label{DT}
	s(x)= 
	\begin{cases} x_{t+1}= A x_{t}+  v_{t} & \\
	y_{t}= C x_{t}+  w_{t} &  
	\end{cases} 
	\end{eqnarray}
	%
	
	%
	
\begin{figure}[!htbp]
		\centering
		\includegraphics[width=4.75cm , height= 8cm]{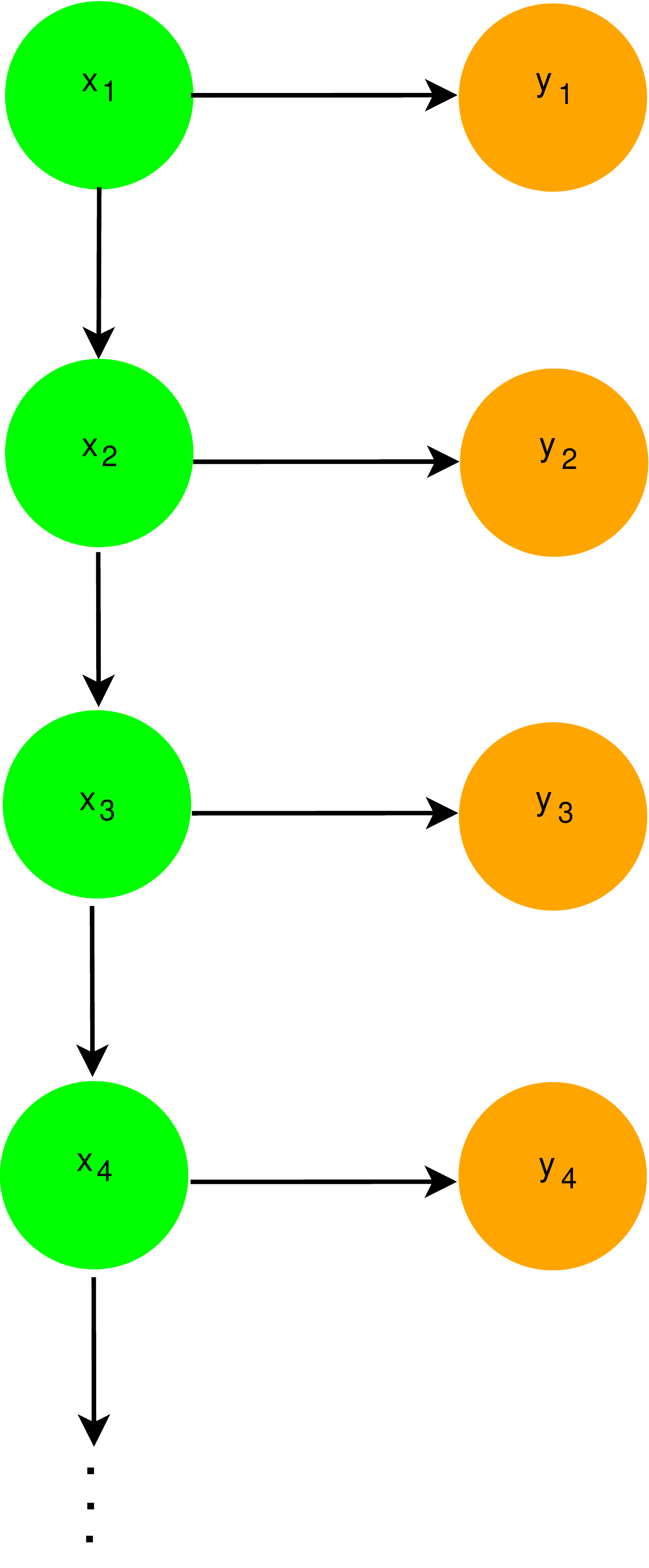}
		\caption{ \label{Fig2} Graphical model for dynamic
			texture DT. $x_t$ and $y_t$ represent the hidden state and the observed video image at the time $t$, respectively.}
	\end{figure}

	%
	Where, the value of the present state $x_t$ represents an essential element for the calculation of the next value of the state variable $x_{t+1}$, and also primordial for the prediction of the present value of the observation process $y_{t}$.  $C \in R^{m}$ is a matrix that contains the main pieces of the video sequence, and the argument $A \in R^{n}$ represents  state-transition matrix. The observation noise $w_{t}$ is  zero mean and Gaussian, with covariance $R$, that is, $w_{t}$ $\sim N(0;R)$, where $R$ $\in$ $R^{m \times m}$. The driving noise process $v_{t}$ is evenly distributed with zero mean and covariance $Q$, that is, $v_{t}$ $\sim N(0;Q)$,  where $Q$ $\in$ $R^{n \times n}$ is a positive-definite $n \times  n$ matrix.  It should be noted
	that a one-dimensional random trajectory in time is defined by each coordinate of the state vector $x_{t}$,  and a value of weighted sum of random trajectories is then assigned to  each pixel, where the coefficients of  weights are included in the corresponding row of $C$.  The
	dynamic texture is completely represented as a graphical model in
	Fig. \ref{Fig2}.
	%

	\section{Proposed unsupervised learning consensus model (ULCM)}
	\label{Method}
\begin{figure*}
			\centering
			\includegraphics[width=15cm, height=22cm]{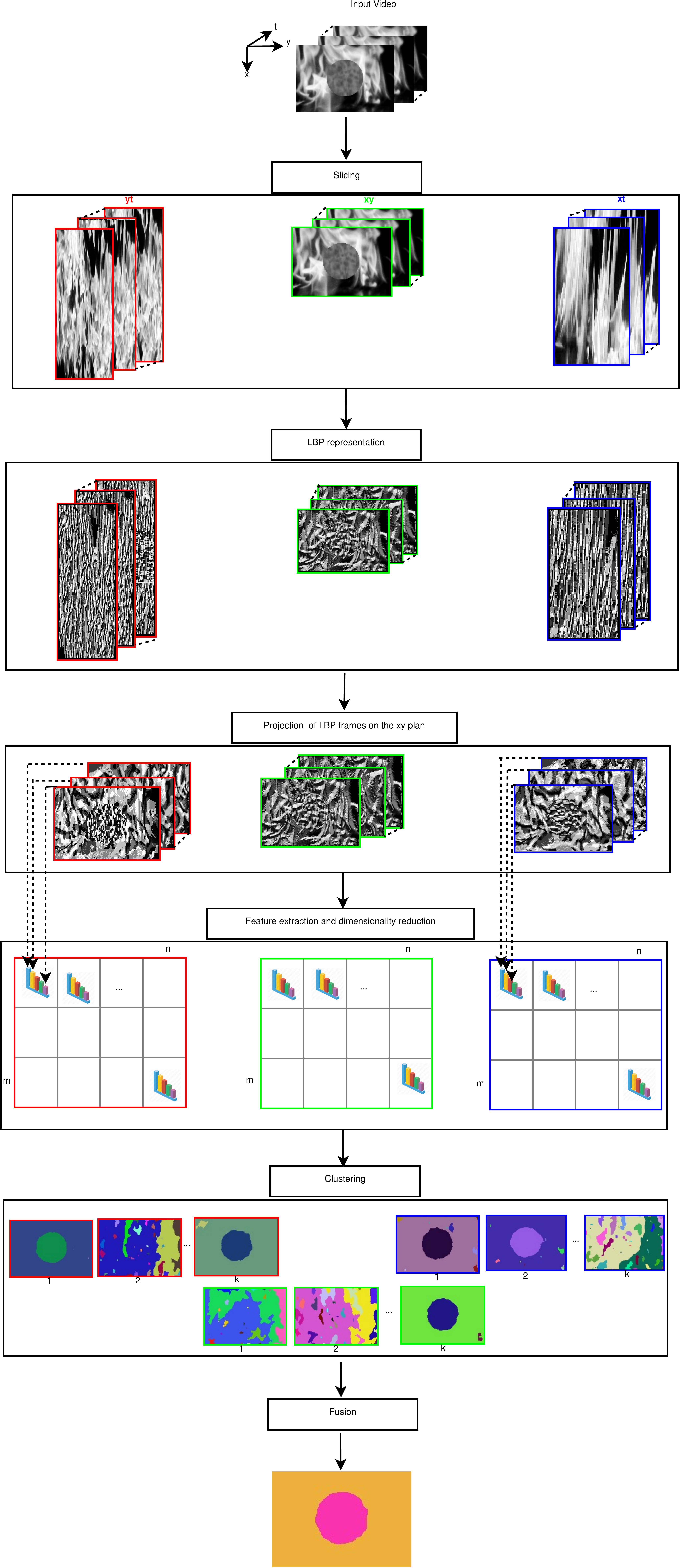}
			\caption{ \label{Fig3}  Proposed system overview.} 
		\end{figure*}

	The method presented here is automatic, straightforward, and consists of five steps, as mentioned in our preliminary work
	\cite{Khelifi2018}.
	In the first step of our method, an ensemble of images 
	is typically obtained by slicing through the video cube (i.e. dynamic texture data). In the second stage, a procedure of feature extraction is proposed and performed for individual images.  In the third stage, for each extracted local histogram associated with each pixel, a variable stochastic dimensionality reduction method using different seeds is performed.  Then, a clustering technique is employed to generate an ensemble of primary segmentations. Once these steps have been completed, in the fourth phase, an energy-based fusion scheme is performed across the ensemble of segmentation maps by iteratively optimizing a deterministic gradient-based optimization algorithm. The pseudo-code of the proposed method is 
	outlined in Fig. \ref{Fig3}.
	
	%

	\subsection{Dynamic Texture Video Clipping}
	\label{Slicing}
	
	In order to fully benefit from the full complementarity of the three
	intrinsic (spatial and temporal) dimensions of our input video
	sequence $V$, and thus to more effectively represent each dynamic
	texture, we perform the following simple slicing operation:
	In addition to the classical slicing process; in which, in the $xy$ spatial plane, we basically produce $w$ equidistant slices that are equally separated in the $t$ time plane  from $V$ corresponding to the $w$ images contained in
	the video sequence, we have added two more clipping processes: First, in the $xt$ time plane, we construct equidistant $h$ slices (or frames) equally separated on the $y$ axis. By this fact, the movement of a line of pixels in time over the length of the video is represented by a slice of the geometric   plane $xt$.  Second, in the time plane $yt$, we produce $m$ equidistant
	frames  equally spaced on the $x$ axis. Concretely,  the evolution of a column of pixels over time along the video sequence is represented by  a slice of the $yt$ plane. Finally, after this slicing stage, we get $h
	\times w \times m$ separate images in three sets (see
	Fig. \ref{Fig2}).

	\subsection{LBP Representation}
	\label{extraction}
	
\begin{figure*}[!t]
		\centering
		\includegraphics[width=13cm, height=13cm]{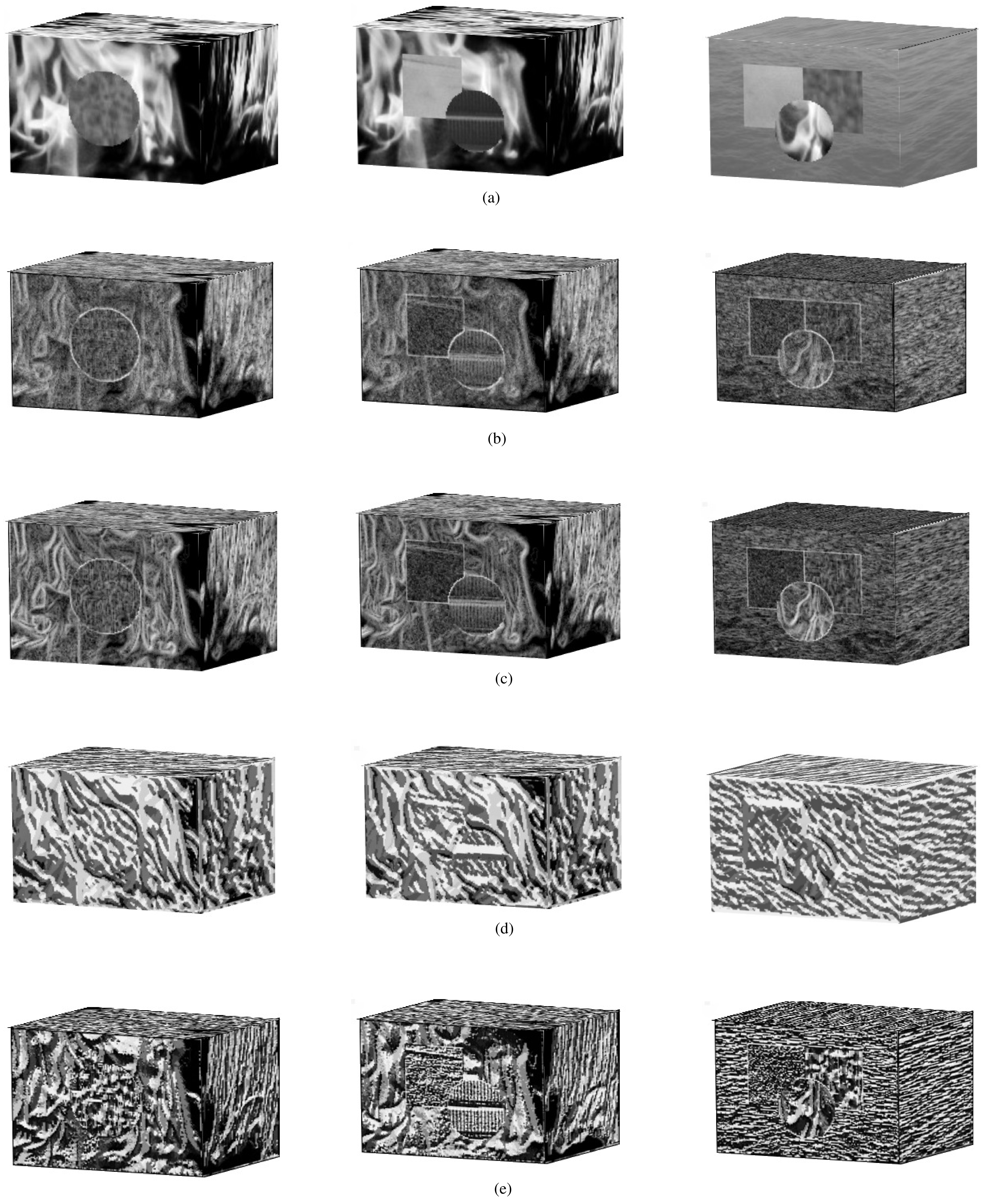}
		\caption{ \label{Fig4} Representation of the input
			video with different texture operator. 
			(a) Original video,
			(b) Histogram of oriented gradients HOG,
			(c) Laplacian operator LAP,
			(d) local phase quantization LPQ,
			(e) local binary pattern LBP.}
	\end{figure*}

	To more efficiently describe the texture, a feature extractor step is adopted by applying a Local binary pattern (LBP) operator to each previously generated frame. (see  Fig. \ref{Fig4}.(e)).  The purpose of using the LBP operator is to describe the statistics of the micro-models within an image (i.e. a frame or slice in our case) by encoding the deviation between  the  central pixel value  and its neighbor's values \cite{Ojala2002}. Formally,  let $q_{c}$ be the value of the center pixel $c$ of a local neighborhood in a gray frame $F$.   Suppose also that  $q_{p}$ $(p=0,...,P-1)$ represent the values of $P$ equidistant pixels uniformly distributed around a circle with radius $R$  forming a circularly symmetric set of neighbors. The coordinates of $q_{p}$ are defined by $(R\sin(\frac{2\pi p}{P}), R \cos(\frac{2\pi p}{P}))$ in the case of coordinates of $q_{c}$ are equal to $(0.0)$.  Particularly, the values of neighbors  that not falling precisely on pixels are obtained by bilinear interpolation.  The LBP descriptor on this pixel ($c$) is given by:
	%
	\begin{eqnarray}
	\label{LBP}
	LBP_{P,R}=\sum_{p=0}^{P-1}s(q_{s}-q_{c})2^{p}, 
	\; s(x)= 
	\begin{cases}1  & ,   x \geq 0\\
	0 &  ,  x < 0
	\end{cases}
	\end{eqnarray}
	%
	
	\subsection{Segmentation Ensemble Generation}
	\label{generation}
	
	By adequately following the steps of our method, a projection process of all the pixels of each LBP generated image onto the $xy$ plane is strongly required (cf. Fig. \ref{Fig3}.(d)). Then,
	for each frame and for each pixel, we estimate, within an overlapping
	square of fixed size  ($Nw= 7$ neighborhood centered around the pixel to
	be classified, a local requantized LBP histogram.
	In the next step,
	we concatenate all local histograms for the individual pixel
	$p^{i}_{(x,y)}$,  each time $t$ to finally create a high-dimensional feature histogram or vector (cf. Fig. \ref{Fig3}.(e)). 
	This high-dimensional histogram encloses a wide range
	of redundant features and masks the potential correlations between data which can make the interpretation of data much
	harder. For that reason, the dimensionality reduction methods \cite{Ayesha2020} may be
	typically used here to avoid   the lack of discrimination, often
	referred to the so-called \textit{curse of dimensionality} problem \footnotemark[3]. In light of this situation, the original features
	should be preprocessed to simplify the high-dimensional histograms
	by transforming it into a low-dimensional structure.  Also, it should be
	noted that, the precision of the segmentation must remain
	satisfactory while minimizing the amount of features to be handled and removing redundant information.  In fact, dimensionality
	reduction is simply the process of projecting the $n$-dimensional
	data onto a subspace of a considerably less lower dimension ($k$)
	that represent a set of principal variables. The commonly used
	approaches include principal component analysis (PCA)
	\cite{Lavanya2018}, multidimensional scaling (MDS) \cite{Marion2019}
	and random projection (RP) \cite{Ella2001} \cite{Wang2019}. In our
	work we resort to  the  random projection (RP) for dimensionality
	reduction for two reasons. Firstly, RP is a much faster and less
	complex (linear complexity) compared to MDS and PCA (quadratic
	complexity). Secondly, RP has the ability to generate, by using different
	seeds, different (and low-dimensional) noisy projected data  which
	will provide the necessary variability to our algorithm which then
	use it efficiently to obtain a final robust segmentation (this will be
	explicit in the following).  Mathematically, in the random
	projection process, the original  data matrix $X$ $[n \times m]$ is
	multiplied by a random projection matrix RP $[m \times k]$ as
	follows:  
	%
	%
	\begin{eqnarray}
	\label{RP}
	X_{red} = \frac{1}{\sqrt{k}} \times X \times RP
	\end{eqnarray}
	%
	where $X_{red}$ is the result of the  projection of the data onto a
	lower $k$ dimensional subspace. Once the dimension reduction step is
	done, in order to generate groups, the clustering algorithm is applied to the different low-dimensional histograms \footnotemark[4] (associated with the different seeds). Hence, we employ the well-known and useful k-means clustering technique \cite{Lloyd82}. We have adopted this choice to assure a reduction in computing time and cost for this important step.

	\footnotetext[3]{
		The curse of dimensionality is a phenomenon that occurs during the analysis of data in high-dimensional spaces. The addition of dimensions extends the points, rendering high-dimensional data highly sparse and uniformly distributed \cite{Debie2019}. This
		sparsity critical for any algorithm for which statistical significance is required.  It is important to note that the organization and retrieval of data are often based on detecting areas where objects are clustered into groups with similar properties (i.e. similar pixels in our case); in high-dimensional data, however, all objects tend to be sparse and disparate in a different way.}
	\footnotetext[4]{
		The size of the final feature vector is 20 times smaller compared to the size of the original high-dimensional vector. 
	}

	\subsection{Fusion Using the Global Consistency Error Criterion}
	\label{GCEMeasure}
	
	Once the set of segmentations is generated, we undertake to merge or fuse all these low quality segmentations into an energy-based fusion model according to the Global Coherence Error (GCE) criterion.
	
	\subsubsection{Global Consistency Error Criterion}
	
	The GCE criterion is a derivative of the so-called local refinement error (LRE), which attempts to quantify the similarity degree in terms of refinement, between two segments \cite{Khelifi2017a}.  According to this perceptual criterion,
	segmentations are supposed to be consistent when they reflect the same image segmented at different levels of detail (or scale)
	\cite{Yang07} \cite{Khelifi16c} or, in other words, when they represent
	a more or less detailed version of the same segmentation. 
	Formally, let suppose that  $n$ is in the number of pixels within the frame $F$ and let $\,
	\Phi_{\mu} \!  = \! \lbrace s^{1}_{\mu},s^{2}_{\mu},\ldots,s^{
		nb_{\mu}}_{\mu}\rbrace$ \& $\Phi_{\mbox{\tiny $\nu$}}=\lbrace
	s^{1}_{\nu },s^{2}_{\nu},\ldots,s^{ nb_{\nu}}_{\nu}\rbrace$ be, two
	segmentation maps of the same frame to be compared, $nb_{\mu}$
	being the number of segments  in $\Phi_{\mbox{\tiny $\mu$}}$ and
	$nb_{\nu}$ the number of segments in $\Phi_{\mbox{\tiny $\nu$}}$. Let
	now $p_i$ be a particular pixel and the couple
	($s^{<p_i\!>}_{\mbox{\tiny $\mu$}}$,$s^{<p_i\!>}_{\mbox{\tiny
			$\nu$}}$) be the two segments containing this pixel, respectively
	in $\Phi_{\mbox{\tiny $\mu$}}$ and $\Phi_{\mbox{\tiny $\nu$}}$. The
	LRE on this pixel $p_i$  is then formulated by:
	%
	\begin{eqnarray}
	\label{LRE}
	\mbox{LRE}(s_{\mbox{\tiny $\mu$}},s_{\mbox{\tiny $\nu$}},p_i) = 
	\frac{|s^{<p_i\!>}_{\mbox{\tiny $\mu$}} \backslash 
		s^{<p_i\!>}_{\mbox{\tiny $\nu$}}|}{|s^{<p_i\!>}_{\mbox{\tiny $\mu$}}|}
	\end{eqnarray}
	%
	where $\backslash$  denotes the algebraic operator of
	difference and $|X|$ the cardinality of the set of pixels $X$. Particularly, a $1$ value signifies that the two regions overlap, inconsistently. Contrarily, an error of $0$ indicates that the pixel is practically in the refinement region. \cite{Martin01}.
	A good way to force all local refinements to go in the same direction
	is to make the LRE metric symmetric. Thereby, , every LRE must be
	measured at least twice, once in each sense, and this simple strategy
	finally leads us to the so-called global coherence error (GCE):
	%
	\begin{eqnarray}
	\label{GCEstar}
	\nonumber
	&& 
	\mbox{GCE}^{\star}(\Phi_{\mbox{\tiny $\mu$}},\Phi_{\mbox{\tiny $\nu$}}) = \\
	&& \frac{1}{2 n}  \, \Biggl \lbrace 
	\sum_{i=1}^{n} \mbox{LRE}(s_{\mbox{\tiny $\mu$}},s_{\mbox{\tiny $\nu$}},p_i) + 
	\sum_{i=1}^{n} \mbox{LRE}(s_{\mbox{\tiny $\nu$}},s_{\mbox{\tiny $\mu$}},p_i)
	\Biggr \rbrace
	\end{eqnarray}
	The GCE$^{\star}$ value lies in the range $[0, 1]$.  A distance of $0$
	indicates a high  similarity (in terms of level of details) between
	the two segmentation maps $\Phi_{\mbox{\tiny $\mu$}}$ and
	$\Phi_{\mbox{\tiny $\nu$}}$. While a distance of $1$ expresses a
	poor consistency or correspondence between the two segmentation maps
	to be  compared.

	\subsubsection{Combination Step}
	
	Now suppose that $\lbrace \Phi_{k} \rbrace_{k \le J} = \lbrace
	\Phi_{1},\Phi_{2},\ldots, \Phi_{J}\rbrace$ represents  the set  of
	$J$ different (weak) segmentation maps to be combined or fused (according
	to the GCE criterion). 
	Let us recall that  $J=3K$, with $K$ being the number of segmentation
	maps which are produced from each set of frames (cf Fig. \ref{Fig3}.(f)).  As already mentioned, our
	ultimate goal is to get the best possible segmentation map $\hat{\Phi}$ of the  video sequence $V$ from this set of multiple  low-cost and weak
	segmentations.
	The fused segmentation result retains all the complementary and redundant information of those weak segmentations \cite{Meher2019}.
	To estimate this refined segmentation result which in fact represents
	a  consensus or a compromise  between these multiple weak
	segmentations, an original and efficient energy-based model framework
	is now proposed to allow us to reconcile (or fuse) these segmentations.
	The main goal of this model is to provide a segmentation map solution as close-as-possible, based on the measured $\mbox{GCE}^{\star}$-distance with respect to all other segmentations $\lbrace \Phi_{k} \rbrace_{k \le J}$. 
	In
	this energy-based framework, if ${\Theta}_{n}$  denotes the set of all feasible segmentations utilizing $n$ pixels, the consensus
	segmentation $\hat{S}_{\overline{\mbox{\tiny GCE}}^{\star}} $ (which is
	optimal according to the $\mbox{GCE}^{\star}$ criterion) is then
	directly given as the minimizer of the following cost functional
	$\overline{\mbox{GCE}}^{\star}$:

	\begin{eqnarray}
	\label{MLFusionModel}
	\hat{\Phi}_{\overline{\mbox{\tiny GCE}}^{\star}} 
	= \arg \min_{\Phi \in { \Theta}_{n}}
	\overline{\mbox{GCE}}^{\star}
	\bigl ( \Phi,\lbrace \Phi_{k} \rbrace_{k \le J} \bigr )
	\end{eqnarray}
	%
	Our fusion  model is thus formulated as an optimization problem
	involving a highly nonlinear cost function.  To optimize this
	nonlinear  function [see Eq (\ref{MLFusionModel})], stochastic
	optimization approaches, for example the simulated annealing
	\cite{Khelifi13}, the genetic algorithm \cite{Mignotte2000} or the
	exploration/selection/estimation (ESE) procedure \cite{Destrempes05}
	may be  successfully exploited.
	These optimizers are assured to find the exact solution, however, with the  drawback of a very high-processing time.
	Another solution that we have followed in this study is a deterministic optimization scheme based on Besag's Iterative Conditional Mode (ICM) method \cite{Besag86} (which is actually also
	equivalent to a Gauss-Seidel based optimization scheme), where each
	label of a pixel is updated one at a time \cite{khelifi17}
	\cite{Khelifi2019}. In the present case, this algorithm has the benefit of
	being simple to implement while also being fast and efficient in terms
	of convergence.

	\section{Experiments and Discussions}
	\label{Experiments}
	\subsection{Experimental Setup}
		\begin{table*} [h]
			\caption{\label{Tab1}
				Performance of the ULCM method compared to other methods on the SynthDB dataset (the higher value of PR index,  is better).}
			\vspace*{1em}
			\renewcommand{\arraystretch}{1.4}
			\centering
			\begin{tabular}{|p{7.5cm}||c|c|c|}
				\hline
				\multirow{3}[0]{6.1em}{\centering ALGORITHMS} &\multicolumn{3}{c|}{  PERFORMANCE (Avg. PR)}\\\cline{2-4}
				& 99 videos& 100 videos & 100 videos \\
				& 2 labels & 3 labels & 4 labels\\
				\hline\hline
				GPCA \cite{Vidal2005} \tiny in \cite{Nguyen2016}   & -0.52- &-0.48- & -0.53-\\\hline
				DTM \cite{Chan2008}                                       & -0.91-& -0.85-&-\textbf{0.86}- \\\hline
				DytexMixCS \cite{Chan2008}                                       & -0.92-& -0.83-&-0.84-\\\hline
				Color (Unsupervised) \cite{Teney2015} & N/A & -0.60- &N/A\\\hline
				AR \cite{Ghoreyshi2006} &66 & N/A & N/A\\\hline
			    AR0 \cite{Ghoreyshi2006} &70 & N/A & N/A\\\hline
				Color + motion (Unsupervised) \cite{Teney2015} & N/A& -0.73-&N/A\\             \hline      
				Color + motion (Learned, logistic regression) \cite{Teney2015} &N/A & -0.78-& N/A\\ \hline
				Color + mouvment (Unsupervised) \cite{Teney2015}  & -0.71-& -0.61-& -0.61- \\\hline
				Color + HoME + mouvment  (Unsupervised) \cite{Teney2015}                                          & -0.86-& -0.80-& -0.74- \\	\hline	
				Ising  \cite{Ghoreyshi2006}  & -0.88-& N/A &  N/A \\\hline
				-\textbf{ULCM}-    & -\textbf{0.95}-& -\textbf{0.86}-& -0.80- \\\hline
			
			\end{tabular}
		\end{table*}
			%
			\vspace{0.5cm}
			\begin{table} [!t]
				\caption{\label{Tab2}
					Performance of the proposed method using different
					fusion criteria on the SynthDB dataset (PR index, higher is
					better).}
				\vspace*{1em}
				\renewcommand{\arraystretch}{1.4}
				\centering
				\begin{tabular}{|p{3cm}||c|c|c|}
					\hline
					\multirow{3}[0]{6.5em}{\centering FUSION CRITERIA} &\multicolumn{3}{c|}{ PERFORMANCE (Avg. PR)}\\\cline{2-4}
					& 99 videos& 100 videos & 100 videos \\
					& 2 labels & 3 labels & 4 labels\\
					\hline\hline              
					-F-measure-  & 0.937& 0.756 &  0.710  \\\hline
					-VoI-  & 0.947& 0.823& 0.763 \\\hline
					-PRI- & 0.919& 0.743 & 0.710 \\\hline
					-GCE- & \textbf{0.953}& \textbf{0.855}& \textbf{0.796} \\
					\hline
				\end{tabular}
			\end{table}
			%
			\begin{table} [!t]
				\caption{\label{Tab3}
					Performance of the ULCM method using different
					texture features on the SynthDB dataset (PR index, higher is
					better).}
				\vspace*{1em}
				\renewcommand{\arraystretch}{1.4}
				\centering
				\begin{tabular}{|p{3cm}||c|c|c|}
					\hline
					\multirow{3}[0]{6.5em}{\centering FEATURES} &\multicolumn{3}{c|}{ PERFORMANCE (Avg. PR)}\\\cline{2-4}
					& 99 videos& 100 videos & 100 videos \\
					& 2 labels & 3 labels & 4 labels\\
					\hline\hline              
					-LAP-  & 0.782& 0.684& 0.674 \\ \hline
					-HOG-  & 0.771& 0.692 & 0.695 \\\hline
					-LPQ- & 0.696 & 0.610 & 0.572 \\\hline
					-OLBP- & \textbf{0.954} & 0.823& \textbf{0.808} \\\hline
					-VLBP- & 0.760 & 0.659& 0.686 \\\hline
					-ELBP- & 0.953& \textbf{0.855}& 0.796 \\
					\hline
				\end{tabular}
			\end{table}
			%

	\textit{Evaluation Datasets.} We have evaluated our model
	quantitatively on SynthDB, a  synthetic video texture
	database\footnotemark[5] \cite{Chan2008}   containing 299 8-bit
	graylevel videos (image size  is 160 $\times$ 110 $\times$ 60
	pixels). The video sequences are divided into three different sets (99 videos with 2 labels, 100 videos with 3 labels, and 100 videos with 4 labels), and a commonly ground truth model is provided for all the three sets. 
	This data set is extremely challenging for two reasons: first, due to the fact that the videos are in grayscale, and also because the textures have a very similar static appearance. In addition, we have evaluated qualitatively  the proposed
	method on the YUP++ \cite{Bida2019} database.

	\textit{Evaluation Metric.} We also rely on the probabilistic Rand
	(PR) index \cite{Unnikrishnan2005} for the evaluation of segmentation
	performance. This metric is widely used in the study of the
	performance of image (sequence) segmentation algorithms.
	More precisely, the PR index metric counts the number of pixel pairs with exactly the same labeling between two image segmentations to be measured.  Mathematically, consider two valid label assignments, an
	automatic segmentation $S_{aut}$ and a manual segmentation (i.e.,
	ground truth) $S_{gt}$ of $N$ pixels $P = {p_1, p_2, . . . p_i,
		. . . , p_N}$ that attribute labels $b_{i}$ and $b^{'}_{i}$
	respectively to pixel $p_i$. The Rand index $R$ can be given as the
	ratio of the number of pairs of pixels with a consistent label
	relationship in $S_{aut}$ and $S_{gt}$. Therefore, the probabilistic rand index (PR) can be considered as follows:
	%
	\begin{eqnarray}
	\label{PR}
	\nonumber
	& R (\tiny S_{aut}, S_{gt})
	= \frac{1}{C^{2}_{N}}\sum_{i,j;i<j}^{n} \bigl [ I(b_{i} = b^{'}_{i} \wedge b_{j} = b^{'}_{j})\\
	& + \; I (b_{i} \neq  b^{'}_{i} \wedge b_{j} \neq  b^{'}_{j}) \bigr ]
	\end{eqnarray}
	%
	where $C^{2}_{N}$ is the number of
	possible unique pairs among N data points and $I$ denotes the identity function. A ``1'' score signifies a good result, while a ``0'' score represents  the worst possible segmentation.
	
	%
	\footnotetext[5]{
		The synthetic video texture database is publicly accessible via this link:
		http://www.svcl.ucsd.edu/projects/motiondytex/}
	%

	%
	\begin{figure*}[h]
		\centering
		\includegraphics[width=18cm]{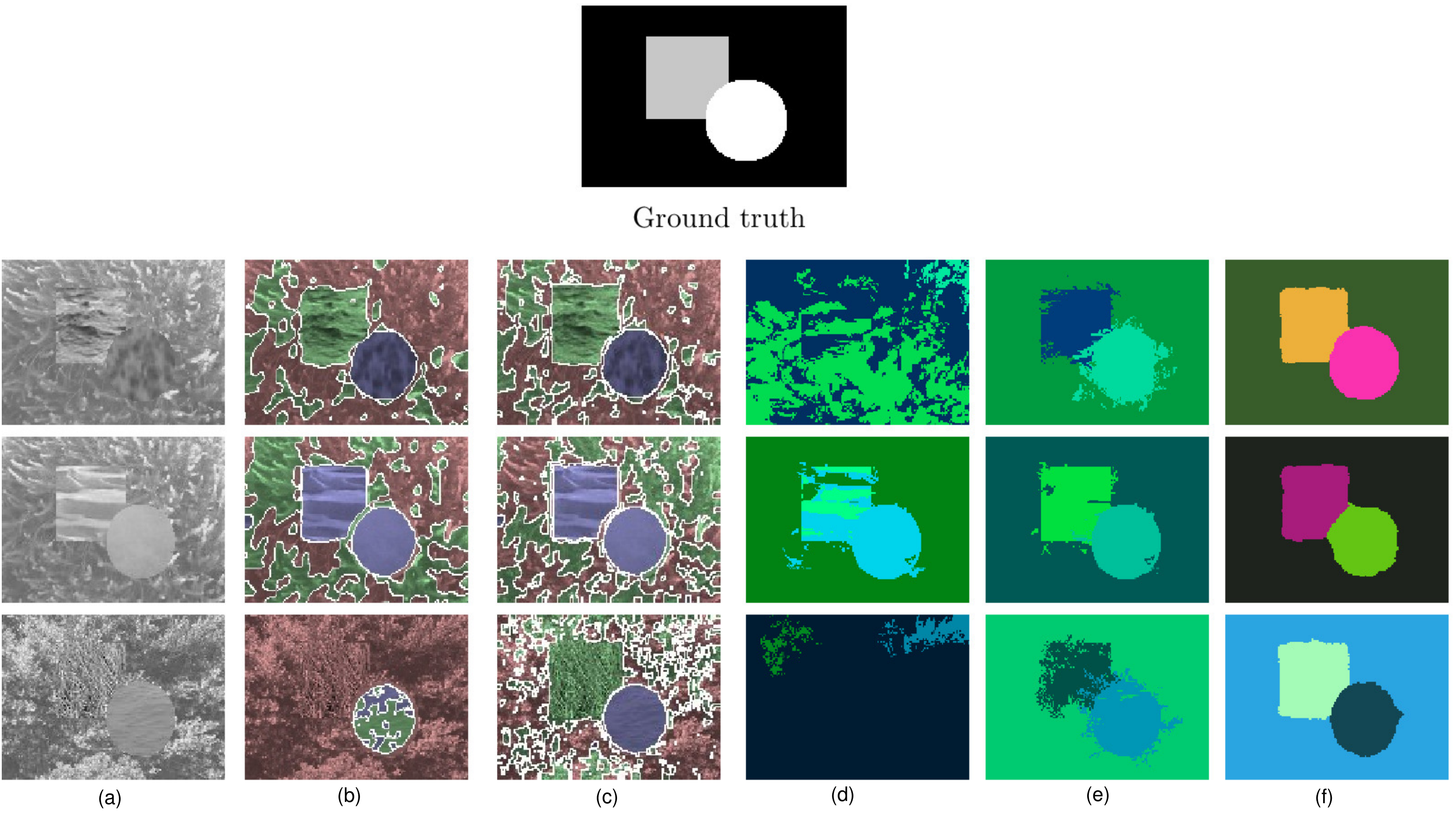}
		\caption{\label{Fig5} Examples of segmentation results produced by the  suggested method on  three videos (with three labels)) of the SynthDB dataset \cite{Chan2008} versus results of other algorithms. (a) Input video, (b) LDT with manual initialization
			\cite{Chan2009},  (c) DTM with contour initialization
			\cite{Chan2008}, (d) Color+motion Unsupervised \cite{Teney2015}
			(f), Color+motion Learned \cite{Teney2015}, (e) Proposed method
			Unsupervised. }
	\end{figure*}

	\subsection{Results and discussions}

\begin{figure*}
		\centering
		\includegraphics[width=18cm]{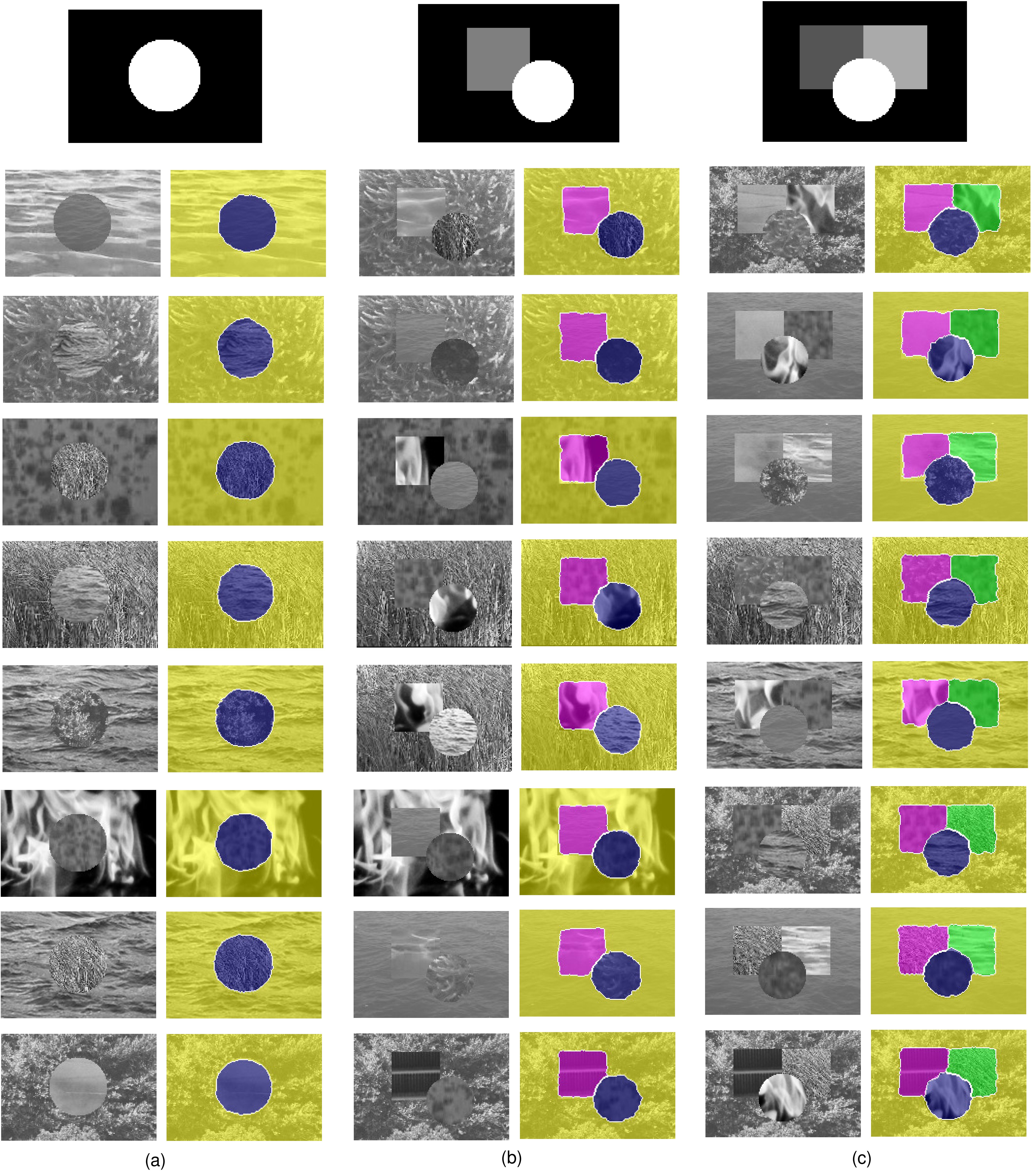}
		\caption{ \label{Fig6} More segmentation results achieved from the SynthDB dataset. (a) one labels, (b) two labels and (c) three
			labels.}
	\end{figure*}

\begin{figure*}
		\centering
		\includegraphics[width=18cm, height=18cm]{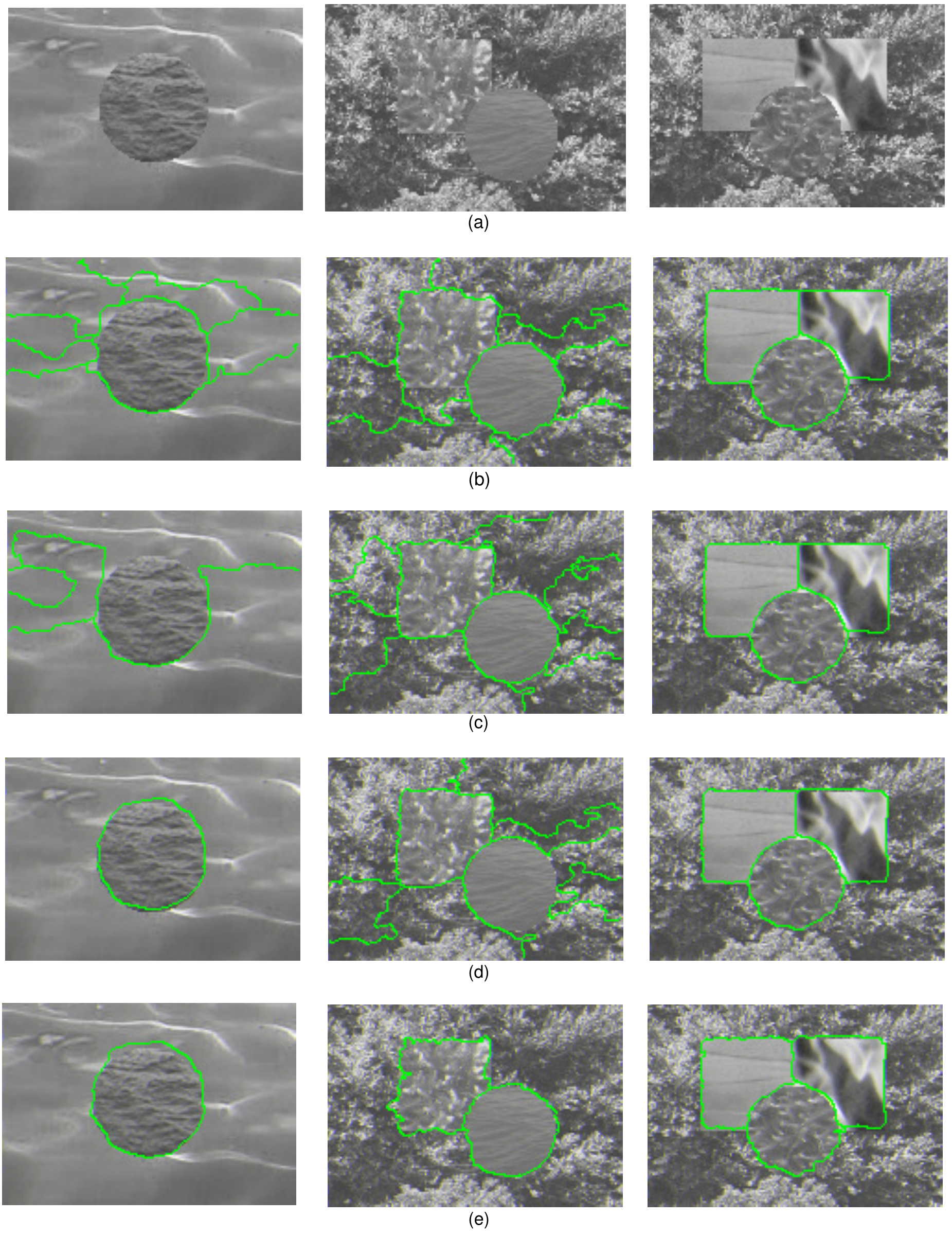}
		\caption{\label{Fig7} Examples of segmentation results
			(generated by the proposed ULCM) of three videos based on different
			fusion criteria. (a) first frame of the video, (b) PRI, (c) VoI,
			(d) F-measure and (e) GCE.}
	\end{figure*}

The performance of the suggested method is compared with the generalized principal
component analysis (GPCA) \cite{Vidal2005}, dynamic texture mixture model (DTM)  \cite{Chan2008}, DytexMixCS \cite{Chan2008}, Ising \cite{Ghoreyshi2006}, AR \cite{Ghoreyshi2006}, AR0 \cite{Ghoreyshi2006}  and  others methods, proposed in \cite{Teney2015},  which are based on different types of features (including color, motion and movement) and metrics.
	Table \ref{Tab1} highlights that the proposed  unsupervised method
	outperforms the other current state-of-the-art methods, even though our
	method has the advantage of not requiring any supervision and/or
	specific initialization step. As a result, we obtain an interesting PR
	score equals 0.953.
	Additionally, to provide a qualitative comparison of the performance of the proposed method versus another set of methods, we present an example of an experiment in Fig. \ref{Fig5}. In this experiment, our model is compared against the layered dynamic textures (LDT) \cite{Chan2009},  the supervised  and
	unsupervised (based learning metric) approaches presented in
	\cite{Teney2015} and the
	dynamic texture model (DTM) \cite{Chan2008}. The result of the presented method, as illustrated in the sixth column, is significantly better as compared to other methods.
	In Fig. \ref{Fig6}  we present additional segmentation
	results obtained  from the SynthDB dataset based on our suggested
	method.
	Results on the complete dataset  are available  publicly on-line in the
	website of the corresponding author at the following http address:
	http://www-etud.iro.umontreal.ca/$\sim$khelifil/ResearchMaterial/consensus-video-seg.html.
	Besides the GCE criterion, We have also tested the effects of using different fusion criteria. Thus, in Table \ref{Tab2}, we report the performances yielded by our algorithm based on the following criteria:  
	
\begin{itemize}
     \item Probabilistic Rand Index
      (PRI), in which agreements and disagreements are weighted
      based on the probability of their occurring by chance \cite{Carpineto2012}.

     \item	
     Variation of information (VoI), in which the information shared between two partitions is measured, in terms of the amount of information that is lost or gained in changing from one clustering to another \cite{KhelifiL2017}.

     \item F-measure which is based on the combination of two complementary measures, namely precision (P) and recall (R) \cite{KhelifiL2017}. 
\end{itemize}
   This test shows that the GCE is the most reliable criteria that yielding the best PR index. In contrast,
	the lower  $PR$ index is achieved based on the PRI criterion with values
	equal  to $0.911$ , $0.743$  and $0.710$, respectively, for videos with
	two, three and four labels. This significant superiority of the GCE criterion is due to its ability to take into account the intrinsically multi-scale nature of image segmentation results.
	As another evaluation test, in
	Fig. \ref{Fig7} we present different  results of segmentation related to  three different videos obtained on the basis of these criteria. Indeed, compared to the PRI, the VOI and the F-measure based results, the GCE
	criterion (in (e))  carries out an improved  qualitative results.  This clearly shows the effectiveness of our choice to use this criterion.
	
	Moreover, in table \ref{Tab3}, we outline  the performance of the
	proposed method using different texture features including;  Laplacian operator (LAP)
     histogram of oriented gradients (HOG),
	local phase quantization (LPQ),
	oriented local binary pattern (OLBP), extended local binary pattern (ELBP) and volume local binary pattern (VLBP).  From this table we can conclude that OLBP and ELBP operator histogram are the features that yield the highest-scoring PR index. One reason for this good results is that OLBP feature combines  together information regarding pixel intensity difference and texture orientation to capture the salient targets.

\begin{figure*}[!t]
		\centering
		\includegraphics[width=18cm,height=21cm]{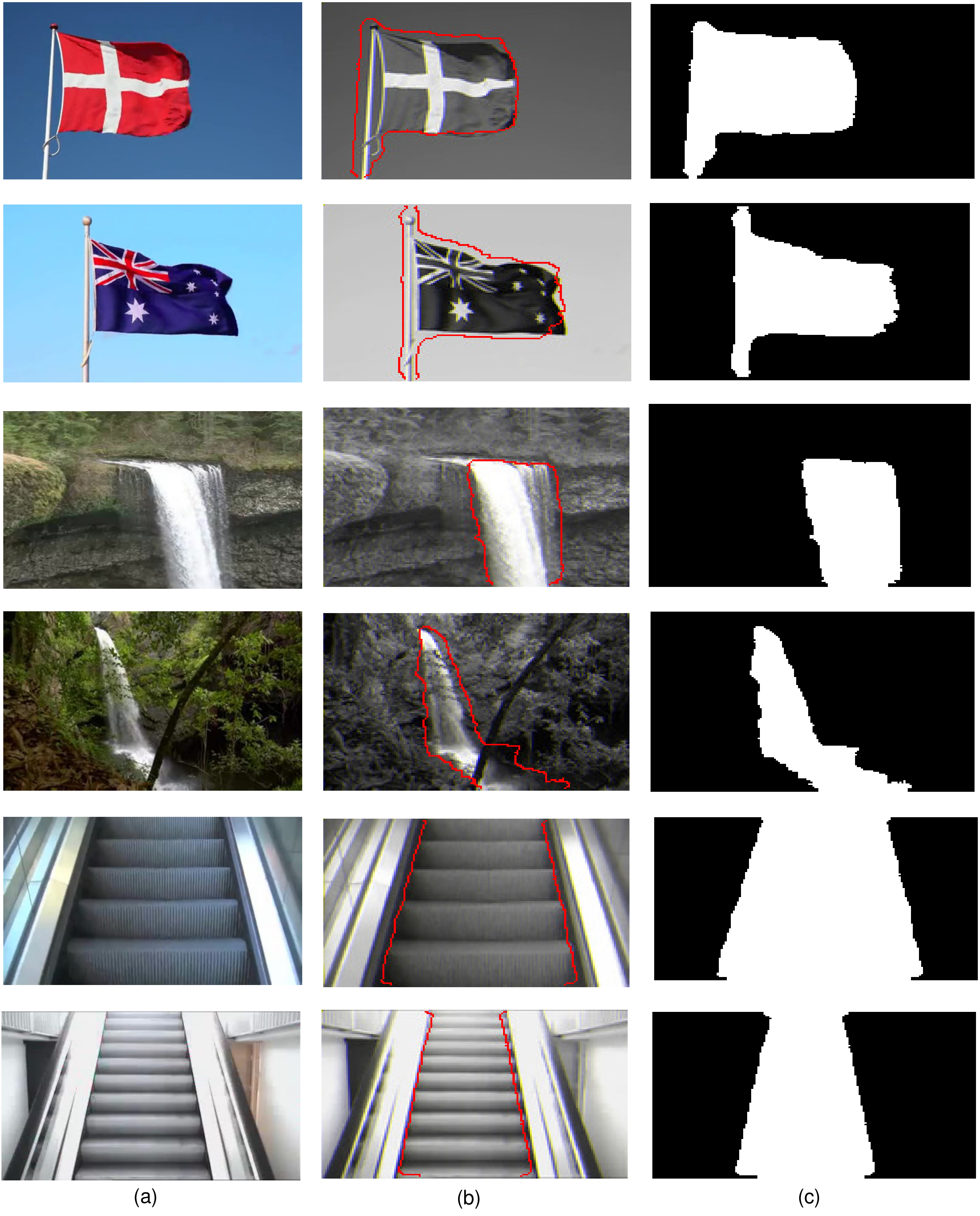}
		\caption{ \label{Fig8} Dynamic textures segmentation
			examples of YUP++ database (WavingFlags, Waterfall and Escalator):
			(a) images, (b) Segmented contours, (c) Segmented masks.}
	\end{figure*}

	With the purpose of testing the robustness of the proposed technique against the variability of dynamic objects, we
	experiment it on the YUP++ database. Thus, in Fig. \ref{Fig8}  we present different segmentation results for complex scenes with waving flags, waterfall and escalator.

	Finally, in Fig.\ref{Fig10} and 
	Fig.\ref{Fig11} we present a  plot of the average PR obtained for each class label (of the SynthDB) and the computing time according to  the dimension of the histogram of features ($k$).
    As can be seen from these figures, the competition time increases considerably as the histogram size increases, and the best average RP value (for the different label categories) is obtained with a value of histogram size in the range of  $[80..120]$.

	\vspace{1ex}
	In summary, our method has the merit of being simple in terms of
	implementation and numerical computation, totally unsupervised while
	being efficient compared to others computationally demanding and  complex
	video segmentation  models that exist in the published literature. In particular, compared to deep learning based models  that require a highly experienced and professional work  to build a large dataset of images specifically annotated  for each object type or class,  our proposed model does not require any human annotations.
	In addition,
	our model remains widely perfectible; either by adding other weak
	segmentations (to be combined) using other interesting (and possibly
	complementary) features  or by using a more efficient fusion criterion
	or distance in our energy-based fusion framework.

		\subsection{Algorithm complexity }

	Table \ref{Tab4} shows in detail the complexity of the main steps of the proposed algorithm.
	Thus, the LBP representation step is characterized by a complexity time of $ O(m \times n \times l \times r \times 3)$, where $m$, $n$, $l$ and $r$ are height, width,  number of frames in the video, number of neighbor points to compute the center pixel, respectively. Here, the multiplication by $3$ is related to our choice in using simultaneous the three plane $xy$, $yt$ and $xt$.
	The complexity of the dimensionality reduction step is equal to $O(m \times n \times l \times k)$, where  $k$ denotes the  dimension of the histogram related to each pixel.
	Moreover, the K-means clustering  step is distinguished by a complexity of
	$O(N \times C \times Iter \times dim)$, where $N$, $C$, $Iter$ and $dim$ are the number of points of each cluster, the number of clusters, the number of iterations and the dimension of each point to be classified, respectively. The last step of fusion  is characterized by a complexity time of $O (Nbreg \times p)$, where $p$ is the pixel number within the image and $Nbreg$ represents the number of regions existing in the set of segmentations (generated by k-means).

			\begin{table} [!t]
				\caption{\label{Tab4}
					Complexity of the difference steps of the ULCM}
				\vspace*{1em}
				\renewcommand{\arraystretch}{1.4}
				\centering
				\begin{tabular}{|p{3cm}||c|}
					\hline
					Steps & Complexity\\
					\hline \hline
					LBP representation	& $O(\times m \times n \times l \times r \times 3)$ \\ \hline
					Dimensionality reduction & $O(m \times n \times l \times k)$ \\ \hline
					Clustering (K-means)	& $O(N \times C \times Iter \times dim)$ \\
					\hline 
					Fusion  & $O(Nbreg \times p )$ \\
					\hline             
				\end{tabular}
			\end{table}
	
	\begin{figure}[!t]
		\centering
		\includegraphics[width=10cm]{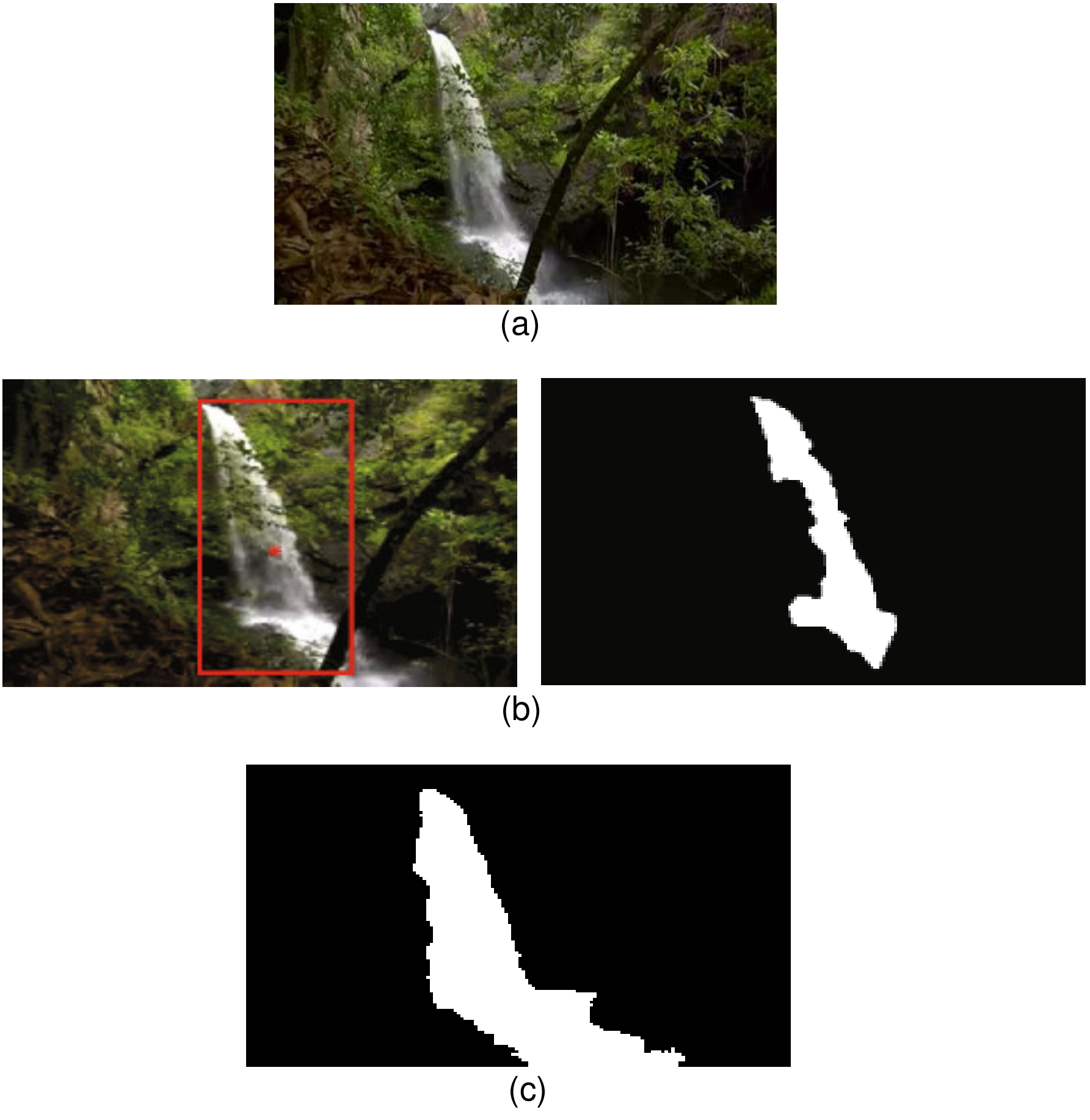}
		\caption{ \label{Fig9} An example of segmentation
			result obtained by our proposed method (without an initialization
			step) from the YUP++ dataset compared to ADAC Algorithm
			\cite{Bida2019} (with an initialization step). (a) input video,
			(b) ADAC , (c) segmentation result based on our method. }
	\end{figure}
	
	\begin{figure}[!t]
		\centering
		\includegraphics[width=10cm]{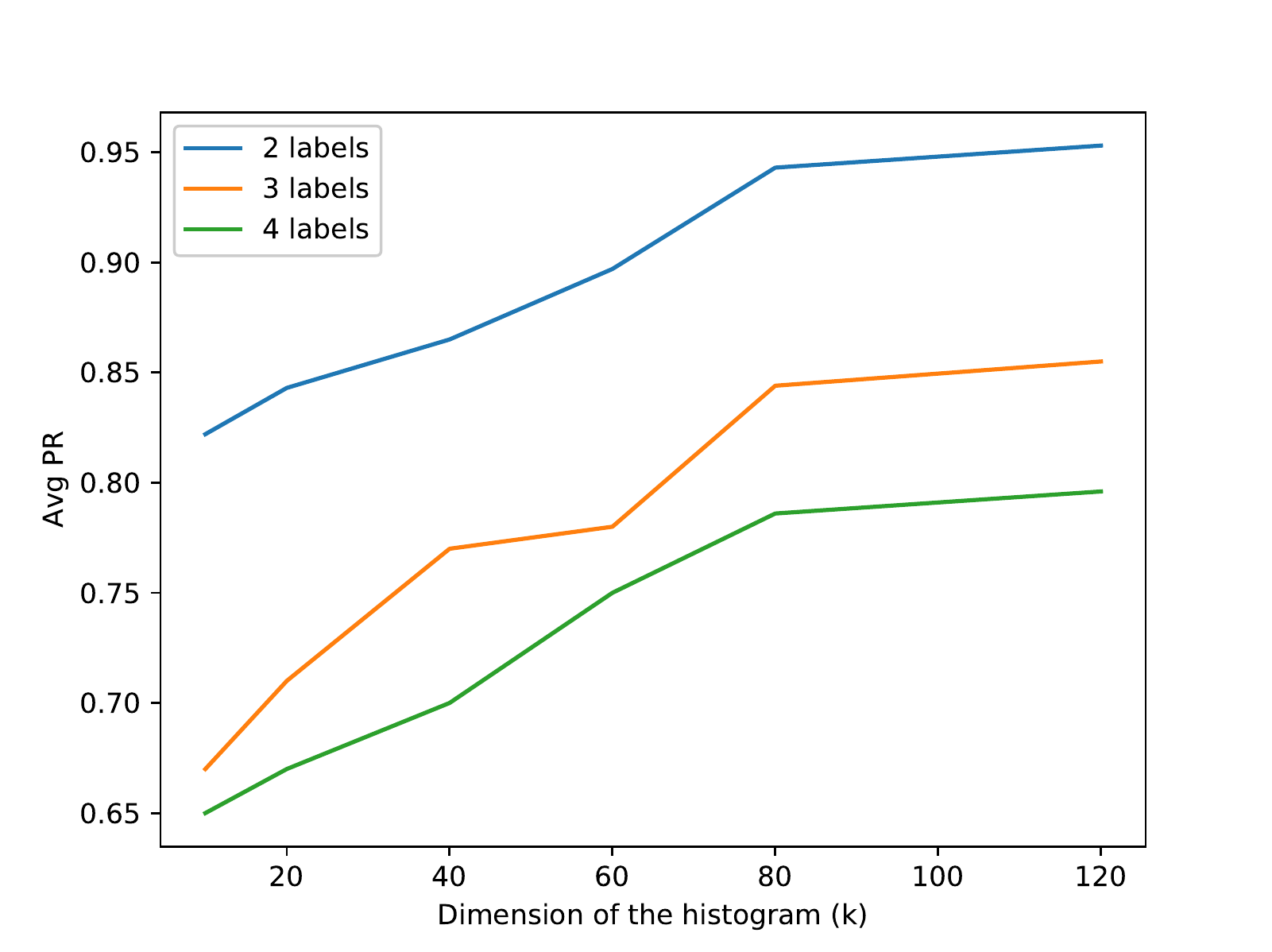}
		\caption{ \label{Fig10} Plot of the average PR obtained for each
			class label (of the SynthDB) as a function of the dimension of the
			histogram of features (k).}
	\end{figure}
	
	\begin{figure}[h]
		\centering
		\includegraphics[width=10cm]{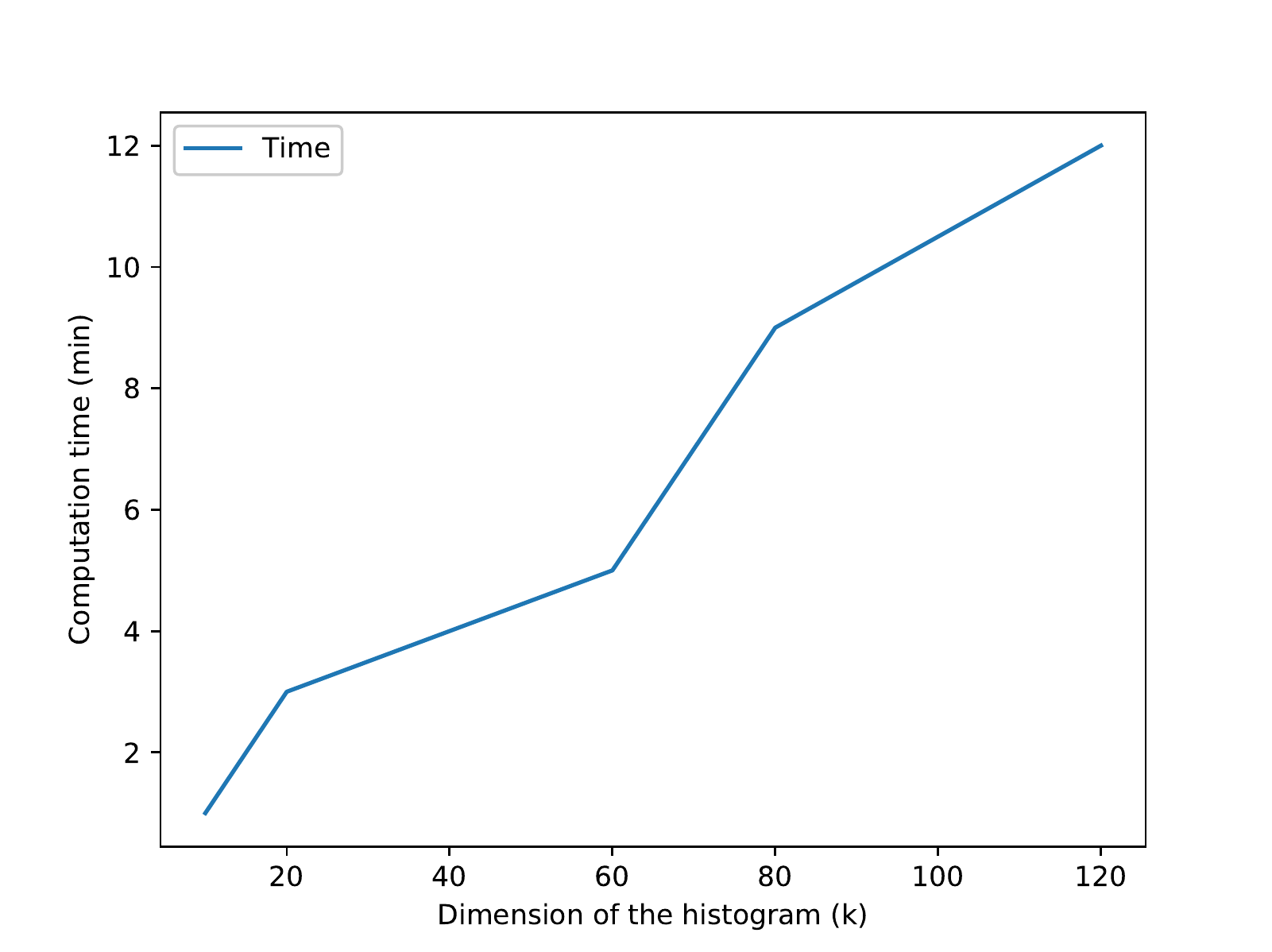}
		\caption{ \label{Fig11} Plot of the computing time  as
			a function of the dimension of the histogram of features (k). }
	\end{figure}
	
	\section{Conclusion}
	\label{Conclusion}
	
	We have presented a new approach to segment  video
	with dynamic textures.  By combining multiple rough (low-quality) region-based
	segmentations of a video and using a new geometric
	criterion, we demonstrated that it is possible
	to achieve a more accurate final segmentation result.
	Experiments show that our model, while being simple, fully unsupervised,
	fast and perfectible, is comparable to the state-of
	the-art methods using supervised or semi-automatic strategy and
	even better than those relying on unsupervised approaches.
	A potential extension of this approach is to incorporate several types of features with the Local Binary Pattern (LBP) to better describe the dynamic texture. Another
	possible extension of this work is to combine other possible
	criteria (variation of information, F-measure and probabilistic rand
	index) to achieve a more reliable result. It is important to mention that the proposed model is adapted to be implemented in parallel or to take full advantage of GPU systems that allow different types of features or criteria to be processed simultaneously.
	

\end{document}